\pdfoutput=1
\documentclass[10pt,twocolumn,letterpaper]{article}
\usepackage{wacv}
\usepackage{times}
\usepackage{epsfig}
\usepackage{graphicx}
\usepackage{amsmath}
\usepackage{amssymb}
\usepackage{lipsum}
\usepackage{enumitem}
\usepackage{multicol}
\usepackage{stfloats}
\usepackage[utf8]{inputenc}
\usepackage[american]{babel}
\usepackage{floatrow}
\usepackage{MnSymbol}
\usepackage{amssymb}
\usepackage{adjustbox}
\usepackage[T1]{fontenc}
\usepackage[font=small,labelfont=bf,tableposition=top]{caption}
\usepackage{float}
\usepackage{bm}
\restylefloat{table}
\restylefloat{figure}

\wacvfinalcopy 

\def\wacvPaperID{546} 

\ifwacvfinal\fi
\begin{document}

\title{MHSAN: Multi-Head Self-Attention Network for Visual Semantic Embedding}

\author{Geondo Park \hspace{2cm} Chihye Han \hspace{2cm} Daeshik Kim\\
Korea Advanced Institute of Science and Technology (KAIST)\\
{\tt\small \{geondopark , ckelseyhan, daeshik\}@kaist.ac.kr}
\and
Wonjun Yoon \\
Lunit Inc.\\
{\tt\small wonjun@lunit.io}
}

\maketitle
\ifwacvfinal\thispagestyle{empty}\fi

\begin{abstract}
Visual-semantic embedding enables various tasks such as image-text retrieval, image captioning, and visual question answering. The key to successful visual-semantic embedding is to express visual and textual data properly by accounting for their intricate relationship. While previous studies have achieved much advance by encoding the visual and textual data into a joint space where similar concepts are closely located, they often represent data by a single vector ignoring the presence of multiple important components in an image or text. Thus, in addition to the joint embedding space, we propose a novel multi-head self-attention network to capture various components of visual and textual data by attending to important parts in data. Our approach achieves the new state-of-the-art results in image-text retrieval tasks on MS-COCO and Flicker30K datasets. Through the visualization of the attention maps that capture distinct semantic components at multiple positions in the image and the text, we demonstrate that our method achieves an effective and interpretable visual-semantic joint space.
\end{abstract}
\section{Introduction}
Various computer vision applications such as image captioning, visual question answering (VQA), and image-text retrieval rely on multi-modal information from images and natural languages. Visual-semantic embedding is to map high-dimensionally represented visual and textual features in the common space where their complex relationship is captured, and it is generally assessed by image-text retrieval. The key to visual-semantic embedding is how accurately input data from each modality are expressed. One popular approach for learning the visual-semantic embedding is to employ a two-path network, which represents visual and textual data separately, using the mono-modal encoders~\cite{Kiros,Faghri,Engilberge}.

Recent works on two-path networks collapse the visual features into a single vector by fully connected layer~\cite{Kiros, Faghri} or selective spatial pooling~\cite{Engilberge} and the textual features using the last hidden state of the recurrent neural network. However, since an image or a sentence consists of a set of sub-components, there can be various descriptions for a single image depending on where the focus is placed within the image, and it is important to propagate information about each component separately when embedding the respective semantics. Previous methods can easily identify the decisive features, but rarely consider the relative importance of each feature and complex semantics.

Motivated by this observation, we propose a multi-head self-attention network for visual-semantic embedding that captures various aspects and components in images and texts. Departing from previous methods, our proposed model generates multiple attention weights focusing on different visuospatial features of the image and semantic features of the sentence in each encoder. Features with a stronger focus are considered more important and have more influence toward the final representation. By explicitly learning `on what point in the image' and `on which word in the sentence' to focus from various perspectives, our model provides more abundant information about visual and textual data for joint embedding. In other words, our model encodes important regions and words separately and weighs the relative importance between them, which is simply collapsed by selective spatial pooling or even single attention.

We evaluate visual-semantic embedding generated by our proposed model on image-text retrieval tasks using the benchmark MS-COCO~\cite{COCO} and Flickr30K~\cite{flickr30K} datasets. The results show that our multi-head attention network establishes the new state-of-the-art record on the two benchmark datasets. Most notably, on Flickr30K~\cite{flickr30K}, our embedding network achieves the relative improvement compared to the current state-of-the-art method~\cite{sco} by 25.4\% in sentence retrieval, and 27.5\% in image retrieval. On MS-COCO~\cite{COCO}, our model also outperforms the current state-of-the-art methods by a large margin. In addition, we perform ablation studies and visualization of the resulting attention maps to demonstrate that the proposed methods of self-attention, multi-head, and diversity regularization each meaningfully contribute to the improved performance.

\textbf{Contribution.} Overall, our main contributions are as follows:
\begin{enumerate}[topsep=0pt,itemsep=-1ex,partopsep=1ex,parsep=1ex]
\item We propose a new visual-semantic embedding network based on the multi-head self-attention module that extracts multiple aspects of semantics in visual and textual data.
\item We demonstrate that our model successfully learns semantically meaningful joint embedding by setting new state-of-the-art records in the image-text retrieval on standard benchmark datasets: MS-COCO and Flickr30K.
\item From the advantage of the interpretability of our model, we validate the effectiveness of multi-head self-attention and diversity regularization.
\item While the method of generating multiple attention is used in the natural language process~\cite{structured}~\cite{EndMN}~\cite{vaswani2017attention}, to the best of our knowledge, this is the first work that generates multiple self-attention maps for image encoding as well as visual-semantic embedding.
\end{enumerate}

\section{Related Work}
\label{sec:relatedwork}
\noindent\textbf{Visual-Semantic Embedding.} Learning visual-semantic embedding aims to map natural images and texts in a common space which captures the complex relations between them and is evaluated by performing image-text retrieval tasks. Recent deep learning based approaches~\cite{Kiros,Faghri,Engilberge, wehrmann2018bidirectional} proposed two-path architectures for learning visual-semantic embedding, which generates image and text representations separately through each mono-modal encoder and connects the two-path with fully connected layers. Kiros \etal~\cite{Kiros} first attempted to create a multi-modal joint embedding space learned using a two-path network with triplet ranking loss. Faghri \etal~\cite{Faghri} suggested advanced triplet ranking loss incorporating hard negative mining and showed a significant improvement in image-text retrieval tasks while keeping the two-path architecture. The aforementioned works used the fully connected layer of the convolutional network trained on the classification task~\cite{imagenet} as the image encoder and recurrent neural network~\cite{LSTM} as the text encoder. For further improvements, Engilberge \etal~\cite{Engilberge} extracted one-dimensional image representation by applying selective spatial pooling to image features to enhance the visual feature encoder. Huang \etal~\cite{sco} proposed learning semantic concepts and order method to enhance the image representation, which is the current state-of-the-art performance in the image-text retrieval tasks. Wehrmann \etal~\cite{wehrmann2018bidirectional} proposed a character-level convolution text encoder using a modified Inception Module~\cite{inception} to improve the textual encoder. Wu \etal~\cite{Univse} improved robustness of visual semantic embedding through structured semantic representations of textual data.\\
With a promising improvement of performance gained by integrating the attention network on many tasks, several works~\cite{DAN,stacked} have tried to perform image-text retrieval tasks with the attention mechanism. DAN~\cite{DAN} applies a cross-attention network, which focuses on the relevant parts of the features between images and texts through multiple step process. Lee \etal~\cite{stacked} proposed the cross-attention network considering relative importance between localized regions of an image and the words that requires the bounding box ground truth which are expensive to pre-train object proposals as used in Fast-RCNN~\cite{fast-rcnn}. In this study, while maintaining two-path, we utilize the self-attention mechanism to focus on important regions and words without object proposals for visual-semantic embedding.\vspace{0.2cm}\\
\textbf{Self-Attention.}
Self-attention networks~\cite{bam}, also called intra-attention networks, are trainable end-to-end, that is, simultaneously train feature extraction and attention generation. There have been several approaches utilizing self-attention to focus on the important parts in order to compute a representation of image and text data~\cite{bam,cbam, bahdanau2014neural,LMattention}. From the motivation to focus on various aspects of the data, several previous works generated attention weights of the same input. Vaswani \etal~\cite{vaswani2017attention} improved machine translation by attending to multiple positions of encoded features. Sukhbaatar \etal~\cite{EndMN} represented the question sentence into a continuous vector by generating attention sequentially for question and answering task. Lin \etal~\cite{structured} have proposed a structured self-attention sentence embedding that represents the sentences with multiple points of attention. MCB~\cite{MCB} achieved a core improvement for visual question answering by their own fusion method and attention mechanism that computes one attention weight from multiple image features and the text feature. In contrast with previous studies, we propose a multi-head self-attention network which generates multiple attention weights that encode the varying localized regions from various perspectives for an image. In our proposed model, self-attention apprehends the various semantics in the image and text for learning visual-semantic embedding.

\begin{figure*}[t]
\begin{center}
\includegraphics[width=1\textwidth]{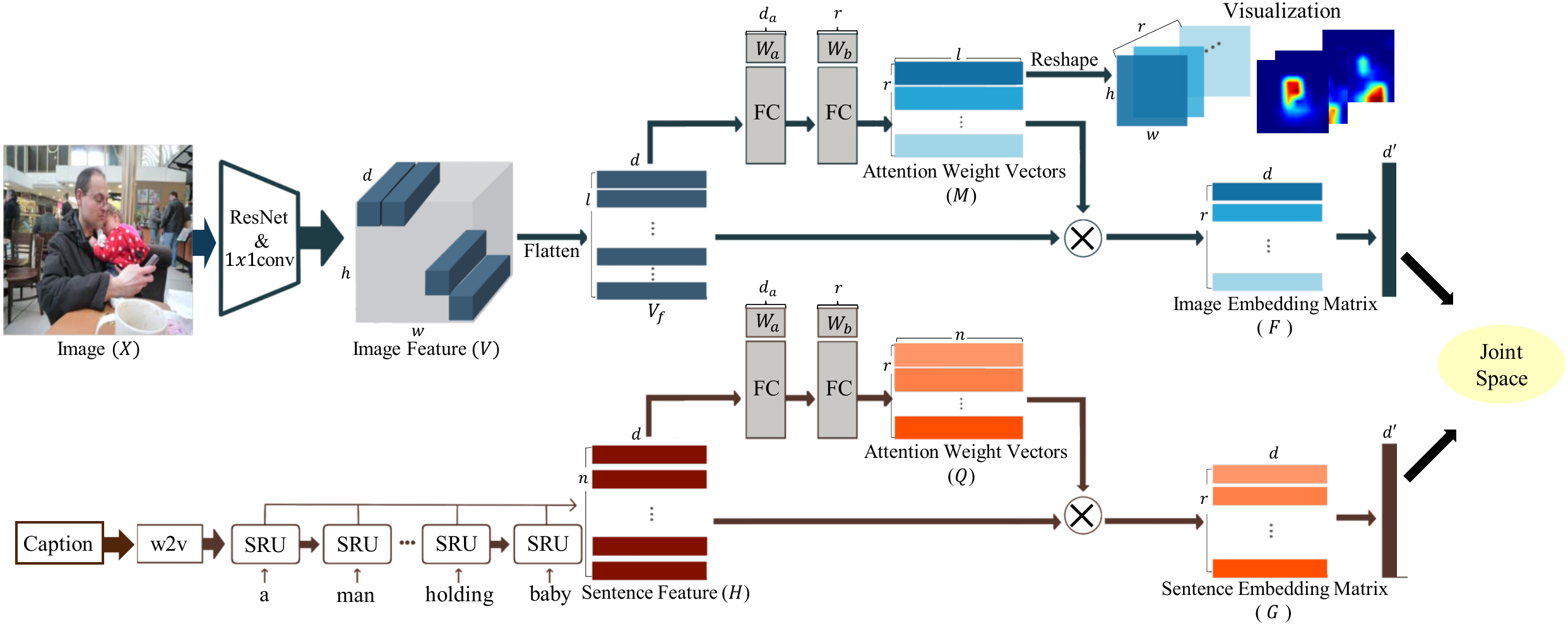}
\end{center}
   \caption{Details of the proposed architecture for learning semantic-visual embedding with multi-head self-attention. For each encoder, features of image$(V)$ and sentence$(H)$ are extracted by ResNet and SRU. On top of the feature extractor, multi-head self-attention network generates attention weight vectors. The final embedding of the image and text is a 2-dimensional matrix.}
\label{fig:model}
\end{figure*}

\raggedbottom
\section{Approach}
\label{sec:approach}
In order to reflect various semantics of visual and textual data in the joint embedding space, we propose multi-head self-attention model generating multiple attention weights, where each weight focuses on the different regions in the image and on the different words in the sentence. Basically, the proposed model follows the two-path network that consists of separate image and text encoders, shown in Figure~\ref{fig:model}. We describe the proposed model in four components: (1) visual feature extractor, (2) textual feature extractor, (3) multi-head self-attention, and (4) contrastive triplet ranking loss with diversity regularization.

\raggedbottom
\subsection{Visual Feature Extractor}
Similar to~\cite{Engilberge}, we use a ResNet-152~\cite{resnet} pre-trained on the ImageNet classification task to extract visual features in the images. In accordance with our purpose, we eliminated all of the fully-connected layers and global average pooling and only used the convolutional layers to extract 3-dimensional feature maps preserving the spatial information of the images. We then superinduce an additional $1\times1$ convolutional layer to perform linear adaptation and to make features in the desired dimension of $d$. Given an image $\bm{X}\in(0,255)^{W \times H \times 3}$ where $W,\,H$ are the width and height of the image, the transformed ResNet-152 outputs the 3-dimensional visual feature of size $\bm{V} \in \mathbb{R}^{w \times h \times d }$ where $(w,h) =(\frac{W}{32}, \frac{H}{32} )$. Then, we flatten the 3-D feature map into a 2-D feature map as 
$\bm{V_{f}} = (v_{1},v_{2},\dotsc,v_{l})$,
where $l= w \times h$ and $v_{1,2,\dots,l}\in \mathbb{R}^{d}$ are features containing spatial information in the images.
\raggedbottom

\subsection{Textual Feature Extractor}
Before extracting feature of the sentences, each sentence is divided into a word unit through the tokenizer and each word is expressed as a vector representation. Given a sentence S with length of $n$ tokens, $\bm{S}=(w_{1}, w_{2},\dots,w_{n})$, $\bm{S} \in \mathbb{R}^{k \times n}$. Here, $w_{i}$ is a $k$-dimensional word vector of a $i$-th word in the sentence and $\bm{S}$ is a set of word vectors that are independent of each other. In this study, we employ $k=620$ word embedding used in~\cite{skipthought} for the word representation. To extract sentence features containing the dependency between each word in a sentence, we use the 4-layer simple recurrent units (SRU)~\cite{sru} with a dropout: $h_{i}=\overrightarrow{SRU}(w_{i},h_{i-1})$ where $h_{i}$ is a $i$-th hidden state of SRU. In contrast to other works which use only the last hidden state $h_{n}$ from recurrent neural networks as a sentence feature, we use all the hidden states as a sentence feature where each hidden state represents the context-aware word in the sentence. Thus, we define the final sentence feature as $\bm{H} = (h_{1},h_{2},\dots,h_{n})$, $ h_{1,2,\dots,n}\in\mathbb{R}^{d}$ where $d$ is the dimension of the hidden states.
\raggedbottom

\subsection{Multi-Head Self-Attention}
We apply the self-attention module on the image and text encoders separately to focus on important spatial regions for image representation and words for sentence representation. Moreover, due to the complex relationship between images and texts, it is difficult to contain the semantic meanings of the image and text within a single vector calculated by a single attention weight.
To solve this problem, we propose the multi-head self-attention mechanism on the image and text encoder to capture the diverse aspects of semantic meaning in the visual and textual data. Another advantage of the multi-head attention architecture is that the representation vectors of an image and sentence retain the important information that could be missed out on single attention which captures globally important information in the whole image scene or sentence. On top of the feature extractor of each modality, the multi-head self-attention module is applied.\\
In case of the image encoder, our goal is to learn a set of attention weight vectors used for refining important spatial features, where each attention weight vector focuses on different subsets of spatial feature. To infer the multiple attention weight vectors, the multi-head self-attention module takes the flattened image features as an input and outputs 2-dimensional attention weight matrix ($\bm{M}$), as illustrated in Figure~\ref{fig:model}. In particular, the flattened features pass through a multi-layer perceptron (MLP) with one hidden state and one activation function. Then, we apply the softmax along the second dimension of output from MLP to ensure the computed each attention weights sum up to 1. Therefore, each row of the attention weight matrix attends to a different part in the image.
In short, the multi-head attention weight matrix is computed as follows:
\begin{equation}\label{eq:all}
\bm{M} = softmax(\bm{W_{a}}ReLU(\bm{W_{b}}\bm{V_{f}^{T}}))
\end{equation}
where $\bm{V_{f}^{T}}\in \mathbb{R}^{d \times l}$, $\bm{W_{b}}\in \mathbb{R}^{d_{a} \times d}$ ,$\bm{W_{a}}\in \mathbb{R}^{r \times d_{a}}$, $\bm{M} \in (0,1)^{r \times l}$, and $d_{a}$ is a hidden state dimension and $r$ is the number of attention weight. In the case of $r=1$, we denote it as single-head self-attention where the model calculates only one attention weight for each image spatial feature.\\
With a set of attention weight vectors, the model generates 2-dimensional image embedding matrix through a linear combination of the flattened image feature,
\begin{equation}\label{eq:finalrep}
\bm{F} = \bm{M}\bm{V_{f}}
\end{equation}
where $\bm{F} \in \mathbb{R}^{r \times d}$ is an image embedding matrix which consists of $r$-vectors. More specifically, given the $\bm{F} = (f_{1},f_{2},\dotsc, f_{r})$, each row vector of $\bm{F}$ is calculated as $f_{r} = \sum\limits_{l=1}^{l}\bm{M_{r,l}}*v_{l}$ where $f_{r} \in \mathbb{R}^{d}$ and $l=w \times h$.\\
The same attention module is applied in the text encoder replacing image features with sentence features. As a result of the text encoder, the sentence embedding matrix ($\bm{G}$) is generated with a set of attention weight vectors ($\bm{Q}$).\\
To map image and text embedding matrix into a joint embedding space, we concatenate $r$-vectors and project into a single vector using additional projection layer with normalization.\\
Further, to visualize the attention maps in Section~\ref{ssec:sub2_ex}, we reshape each attention weight vector into a feature size of $(w, h)$ and resize into the same size as the input image using bi-linear interpolation. For multi-head self-attention, $r$-attention heatmaps where each heatmap focuses on a different local region in an image are provided.

\subsection{Contrastive Triplet Ranking Loss with Diversity Regularization}
We basically use the triplet ranking loss with hard negative ($Loss_{T}$), which has shown a significant improvement on image-text retrieval tasks in the previous studies~\cite{Faghri,Engilberge,stacked}. Since the multi-head attention mechanism can suffer from the redundancy problem where each attention vector focuses on the very similar region, we use the diversity regularization loss ($Loss_{D}$) with a coefficient($\lambda$) employed in~\cite{structured}:
\begin{equation} \label{eq:total}
Loss_{total} = Loss_{T} + \lambda Loss_{D}
\end{equation}
Given a sampled training set of image and text pairs in a batch: $B = {\{(X_{n},Y_{n})\}}_{n=1}^{N}$ where $X_{n}$ is the $n$-th image and $Y_{n}$ is the $n$-th sentence in a batch, the image vector $x_{n}$ denote the image $X_{n}$ and sentence vector $y_{n}$ denote the sentence $Y_{n}$ in the joint embedding space. Then, triplet ranking loss with hard negative is defined as:
\begin{equation}
\begin{aligned}\label{eq:Triplet}
 Loss_{T} = \frac{1}{N}\sum\limits_{n=1}^{N}( \max_{m \in N, m \neq n} [\alpha - cos(x_{n}, y_{n}) + cos(x_{n}, y^{-}_{m})]_{+}
 \\+ \max_{m \in N, m \neq n} [\alpha - cos(y_{n}, x_{n})  + cos(y_{n}, x^{-}_{m})]_{+})
\end{aligned}
\end{equation}
where $y^{-}_{m}$ , $x^{-}_{m}$ are negative instances, $(x_{n},y_{n})$ is a positive pair in a batch, $\alpha$ is the margin and $[t]_{+} = max(0,t)$. The contrastive triplet ranking loss encourages the matched embedding pairs to be closer at least margin $\alpha$ apart than mismatched embedding pairs in the joint embedding space. Following the hard negatives mining strategy, the loss function focuses only on a single contrastive example that has the highest similarity score for image and text in the batch while not similar in meaning.\\ 
To ensure the diversity of multiple attention weight vectors in each encoder, we add the diversity regularization loss defined as:
\begin{equation} \label{eq:regularization}
Loss_{D} = \|(\bm{M}\bm{M^{T}} - \bm{I})\|_{F}^{2} + \|(\bm{Q}\bm{Q^{T}} - \bm{I})\|_{F}^{2}
\end{equation}
where $\bm{M}$ and $\bm{Q}$ is the R-multiple attention weight matrix from the each image and text encoder, $\bm{I} \in \mathbb{R}^{r \times r}$ is the identity matrix and $\|\cdot\|_{F}$ denotes the squared frobenius norm. Minimizing the regularization term with triplet ranking loss, the matrices $\bm{M}$ and $\bm{Q}$ become sparse so that each of the attention vector focuses on the different region and word in each encoder. We show the effect of the diversity regularization in Section~\ref{ssec:sub3_ex}.
\subsection{Implementation Details}
\noindent\textbf{Image Encoder} First, we take a random rectangular crop from an image and resize the image into a fixed size of (256 $\times$ 256) for training images. Then, we use ResNet-152 and 1$\times$1 convolution where the output dimension ($d$) is set to 2400. We set a hidden state of MLP, $d_{a}$, to 350 and the joint embedding dimension, $d'$ to 2400. We use 0.5 probability of dropout on the MLP of self-attention mechanism and the projection layer for the joint embedding.\vspace{0.2cm}\\
\textbf{Text Encoder}
We use tanh and 0.25 dropout in the SRU text encoder. The self-attention mechanism and the projection layer is set the same as the image encoder. Unlike the self-attention mechanism in the image encoder, we use tanh instead of relu between MLP.\\
\textbf{Training Process}
We set the margin of triplet ranking loss, $\alpha$, to 0.2 and the coefficient of diversity regularization to 0.1. We use the Adam optimizer for the whole training. We first update only the last projection layer of the image encoder for the 4 epochs with 0.005 initial learning rate. Then, we update the SRU for the 15 epochs with 0.1 decayed learning rate. Finally we train the rest of the parameters in the image encoder together for the 40 epochs with 0.1 decayed learning rate.
\section{Experiment}
\label{sec:experiment}
\begin{table}[t]
\begin{center}
\begin{adjustbox}{max width=1.0\textwidth}
\begin{tabular}{l|ccc|ccc}
\hline
Method &\multicolumn{3}{c|}{Sentence Retrieval} & \multicolumn{3}{c}{Image Retrieval}\\
&R@1&R@5&R@10&R@1&R@5&R@10\\
\hline
&\multicolumn{6}{c}{1K Test Images}\\
\hline
Order-embeddings~\cite{order_emb} &46.7 &- &88.9 &37.9 &- &85.9\\
Embedding network~\cite{embeddingnetwork}  &50.4  &79.3  &89.4  &39.8  &75.3  &86.6\\
2-Way Net~\cite{twowaynet}  &55.8  &75.2  &-  &39.7  &63.3  &-\\
CHAIN-VSE~\cite{wehrmann2018bidirectional}  &61.2  &89.5  &95.8  &46.6  &81.9  &90.92\\
NAN~\cite{sco}  &61.3  &87.9  &95.4  &47.0  &80.8  &90.1\\
UniVSE~\cite{Univse}&64.3&89.2&94.8&48.3&81.7&91.2\\
VSE++~\cite{Faghri}  &64.6  &90.0  &95.7  &52.0  &84.3  &92.0\\
DPC~\cite{DPC}  &65.6  &89.8  &95.5  &47.1  &79.9  &90.0\\
GXN~\cite{GXN}  &68.5  &-  &97.9  &56.6  &-  &94.5\\
Engilberg \etal~\cite{Engilberge}  &69.8  &91.9  &96.6  &55.9  &86.9  &94.0\\
SCO~\cite{sco}  &69.9  &92.9  &97.5  &56.7  &87.5  &\textbf{94.8}\\
\hline
Ours (1-head)&72.4&93.7&97.5&57.5&88.0&94.4\\
Ours (5-head)&72.6&93.7&97.6&58.5&88.0&94.5\\
Ours (10-head)&\textbf{73.5}&\textbf{94.2}&\textbf{98.1}&\textbf{59.1}&\textbf{88.5}&\textbf{94.8}\\
\hline
&\multicolumn{6}{c}{5K Test Images}\\
\hline
Order-embeddings~\cite{order_emb} &23.3 &- &84.7 &31.7 &- &74.6\\
VSE++~\cite{Faghri} &41.3 &71.1 &81.2 &30.3 &59.4 &72.4\\
GXN~\cite{GXN}  &42.0  &-  &84.7  &31.7  &-  &74.6\\
SCO~\cite{sco}  &42.8  &72.3  &83.0  &33.1  &62.9  &75.5\\
\hline
Ours (10-head)&\textbf{49.3}&\textbf{78.3}&\textbf{87.1}&\textbf{36.2}&\textbf{67.2}&\textbf{78.7}\\
\hline
\end{tabular}
\end{adjustbox}
\end{center}
\vspace{-0.3cm}
\caption{Results of experiments on MS-COCO, the best in bold.}
\label{table:1}
\end{table}
We first evaluate the quantitative result of our cross-modal retrieval model on the standard image-text aligned datasets, MS-COCO and Flickr30K, and compare our models with recent other methods. Then, we perform additional experiments to demonstrate that the self-attention, multi-head, diversity regularization on each encoder contribute to respective performance improvement. Additionally, through visualization of the attention maps, we qualitatively show how our model works and what our model has propagated into the joint embedding space. Our pytorch implementation is available as open source.\footnote{Github: https://github.com/GeondoPark/MHSAN}\footnotetext[2]{We use the bidirectional SRU in the text extractor for this results.}\vspace{0.2cm}\\
\textbf{Dataset and Evaluation Metric}. MS-COCO and Flicker30K datasets consist of 123,287, 31,783 images respectively and 5 captions are annotated for each image. We follow training/validation/test splits used in~\cite{karpathy2015deep}: 1)MS-COCO, 113,287 training, 5,000 validation, 5,000 test images. 2)Flickr30K, 29,783 training, 1,000 validation, 1,000 test images. For comparing the performance of our model with state-of-the-art methods in image-text retrieval, we report the same evaluation metric with previous works~\cite{Faghri, Engilberge, stacked, wehrmann2018bidirectional, sco}. Recall@K of the text retrieval is the percentage of the correct text being ranked within the top-K retrieved results, and vice versa. For MS-COCO dataset, we additionally report averaging R@K results, which are performed 5 times on 1,000-image subsets of the test set and testing on the full 5,000 test images.
\subsection{Comparison with the state-of-the-art}
\noindent\textbf{MS-COCO Retrieval Task}. Table~\ref{table:1} shows the quantitative results of our model on MS-COCO dataset. We denote our models with the number of attention weight vectors (\#-head), which is the same in the image and text encoders. Comparing with recently proposed methods, our model with 10-head attention network achieves the new state-of-the-art performance on MS-COCO datasets.
On the 1K test set experiment, our models outperform SCO~\cite{sco}, the current state-of-the-art based on the visual-semantic embedding method, by 5.2\% in sentence retrieval (R@1) and by 4.2\% in image retrieval (R@1). Also, our best results on 5k test set experiment show 15.2\% and 9.4\% relative improvement in sentence retrieval (R@1) and in image retrieval (R@1), respectively.\vspace{0.2cm}\\
\textbf{Flickr30K Retrieval Task}. We also report the experimental results on Flickr30K, which is a relatively smaller dataset in Table~\ref{table:2}. The network architecture and dimension of the joint embedding are the same as those used for MS-COCO. Compared to the recently proposed methods, our model achieves the best performance across all measures with a large gap. Our model with 10-head relatively improves 25.4\% on sentence retrieval (R@1) and 27.5\% on image retrieval (R@1) compared with the current state-of-the-art score from SCO~\cite{sco}. This result also demonstrates that our model works well for small dataset, which can cause overfitting emerged in the previous two-path architectures~\cite{Engilberge, wehrmann2018bidirectional}.
\begin{table}
\begin{center}
\begin{adjustbox}{max width=1.0\textwidth}
\begin{tabular}{l|ccc|ccc}
\hline
Method &\multicolumn{3}{c|}{Sentence Retrieval} & \multicolumn{3}{c}{Image Retrieval}\\
&R@1&R@5&R@10&R@1&R@5&R@10\\
\hline
2-Way Net~\cite{twowaynet}&49.8&67.5&-&36.0&55.6&-\\
VSE++~\cite{Faghri}  &52.9  &80.5  &87.2  &39.6  &70.1  &79.5\\
DAN~\cite{DAN}  &55.0  &81.8  &89.0  &39.4  &69.2  &79.1\\
NAN~\cite{NAN}  &55.1  &80.3  &89.6  &39.4  &68.8  &79.9\\
DPC~\cite{DPC}  &55.6  &81.9  &89.5  &39.1  &69.2  &80.9\\
SCO~\cite{sco}  &55.5  &82.0  &89.3  &41.1  &70.5  &80.1\\
\hline
Ours (1-head)  &66.2  &89.8  &93.9  &49.0  &79.0  &86.5\\
Ours (5-head)  &67.5  &89.6  &94.8  &51.8  &80.0  &87.8\\
Ours (10-head)\footnotemark[2]&\textbf{69.6}&\textbf{92.2}&\textbf{95.8}&\textbf{52.4}&\textbf{80.7}&\textbf{88.6}\\
\hline
\end{tabular}
\end{adjustbox}
\end{center}
\vspace{-0.3cm}
\caption{Results of experiments on Flickr30K, the best in bold.}
\label{table:2}
\end{table}
\subsection{Comparison of Self-Attention and Spatial Pooling}\label{ssec:sub1_ex}
\begin{figure*}[t]
\begin{center}
\includegraphics[width=1.0\textwidth]{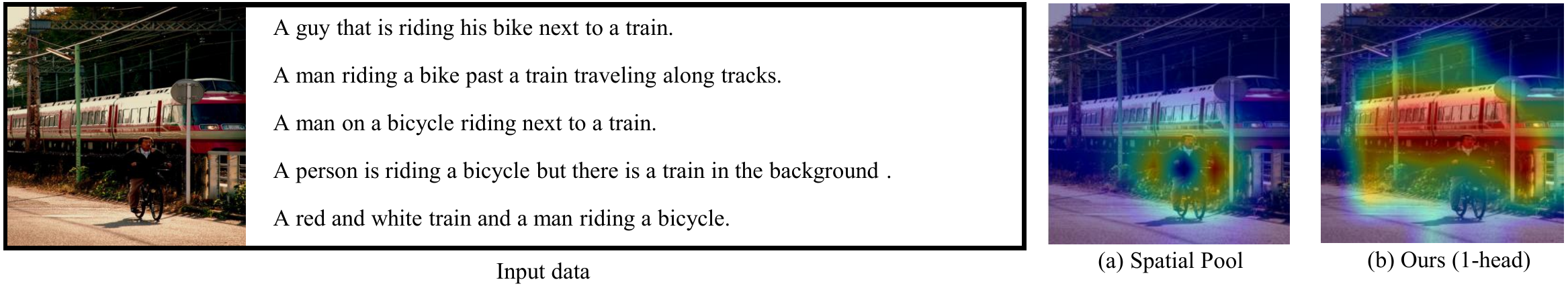}
\end{center}
\vspace{-0.3cm}
   \caption{Comparison of single-head (self-attention) with spatial pooling (without attention)}
\label{fig:spatial1hop}
\end{figure*}
\begin{table*}[h]
\begin{center}
\begin{adjustbox}{max width=0.9\textwidth}
\begin{tabular}{l|l|c|c}
\hline
Image Encoder&Text Encoder&Sentence Retrieval(R@1)&Image Retrieval(R@1)\\
\hline
\hline
W/O attention ~\cite{Engilberge}&W/O attention ~\cite{Engilberge}&69.8&55.9\\
\hline
Our image encoder(1-head)&W/O attention ~\cite{Engilberge}&\textbf{70.1}&55.8\\
\hline
W/O attention ~\cite{Engilberge}&Our text encoder(1-head)&69.9&\textbf{56.2}\\
\hline
\end{tabular}
\end{adjustbox}
\end{center}
\vspace{-0.3cm}
\caption{Effect of self-attention (1-head) in each encoder. We denote the image encoder using selective spatial pooling and the text encoder using the last hidden state which are used in Engilberg \etal~\cite{Engilberge} as W/O attention. We report only R@1 for effective comparison.}
\label{table:3}
\end{table*}
\begin{table*}[b]
\begin{center}
\begin{adjustbox}{max width=0.9\textwidth}
\begin{tabular}{l|l|c|c}
\hline
Image Encoder&Text Encoder&Sentence Retrieval(R@1)&Image Retrieval(R@1)\\
\hline
\hline
Our image encoder(1-head)&W/O attention&70.1&55.8\\
Our image encoder(5-head)&W/O attention&70.9&56.2\\
Our image encoder(10-head)&W/O attention&\textbf{71.1}&\textbf{57.1}\\
\hline
W/O attention&Our text encoder(1-head)&69.9&56.2\\
W/O attention&Our text encoder(5-head)&71.2&57.4\\
W/O attention&Our text encoder(10-head)&\textbf{71.7}&\textbf{58.0}\\
\hline
Our image encoder(1-head)&Our text encoder(1-head)&72.4&57.5\\
Our image encoder(10-head)& Our text encoder(10-head)&\textbf{73.5}&\textbf{59.1}\\
\hline
\end{tabular}
\end{adjustbox}
\end{center}
\vspace{-0.3cm}
\caption{Effect of the number of heads in each encoder. We denote the image encoder using selective spatial pooling and the text encoder using the last hidden state which are used in Engilberg \etal~\cite{Engilberge} as W/O attention. We report only R@1 for effective comparison.}
\label{table:4}
\end{table*}
In this subsection, we show that the single-head self-attention improved the performance of retrieval tasks compared to the selective spatial pooling~\cite{Engilberge} that did not use the attention mechanism. For this experiment, we use the MS-COCO dataset. To see the effect of the single-head self-attention mechanism at each encoder, we apply the single-head attention to each encoder separately. Table~\ref{table:3} shows that our model with a single-head attention at each encoder slightly improves the performance in the image-text retrieval task.\\
\textbf{Visualization.} We validate qualitatively why the self-attention performs better than the network using selective spatial pooling~\cite{Engilberge} through visualization of the attention heatmaps. In Figure~\ref{fig:spatial1hop}, we present a comparison of attention heatmap computed by our model with the single-head and the selective spatial pooling heatmaps. For visualizing which region is considered for embedding in the selective spatial pooling, we employ the segmentation method proposed in~\cite{Engilberge} substituting a phrase with a full sentence. We observe that self-attention with a single-head captures the missing parts on the selective spatial pooling, e.g. the train contained in descriptions of the image as shown in Figure~\ref{fig:spatial1hop}. Through the visualization, we demonstrate that our model is more suitable for visual semantic embedding, where the encoding of rich details is necessary.
\begin{figure*}[t]
\begin{center}
\includegraphics[width=0.7\textwidth]{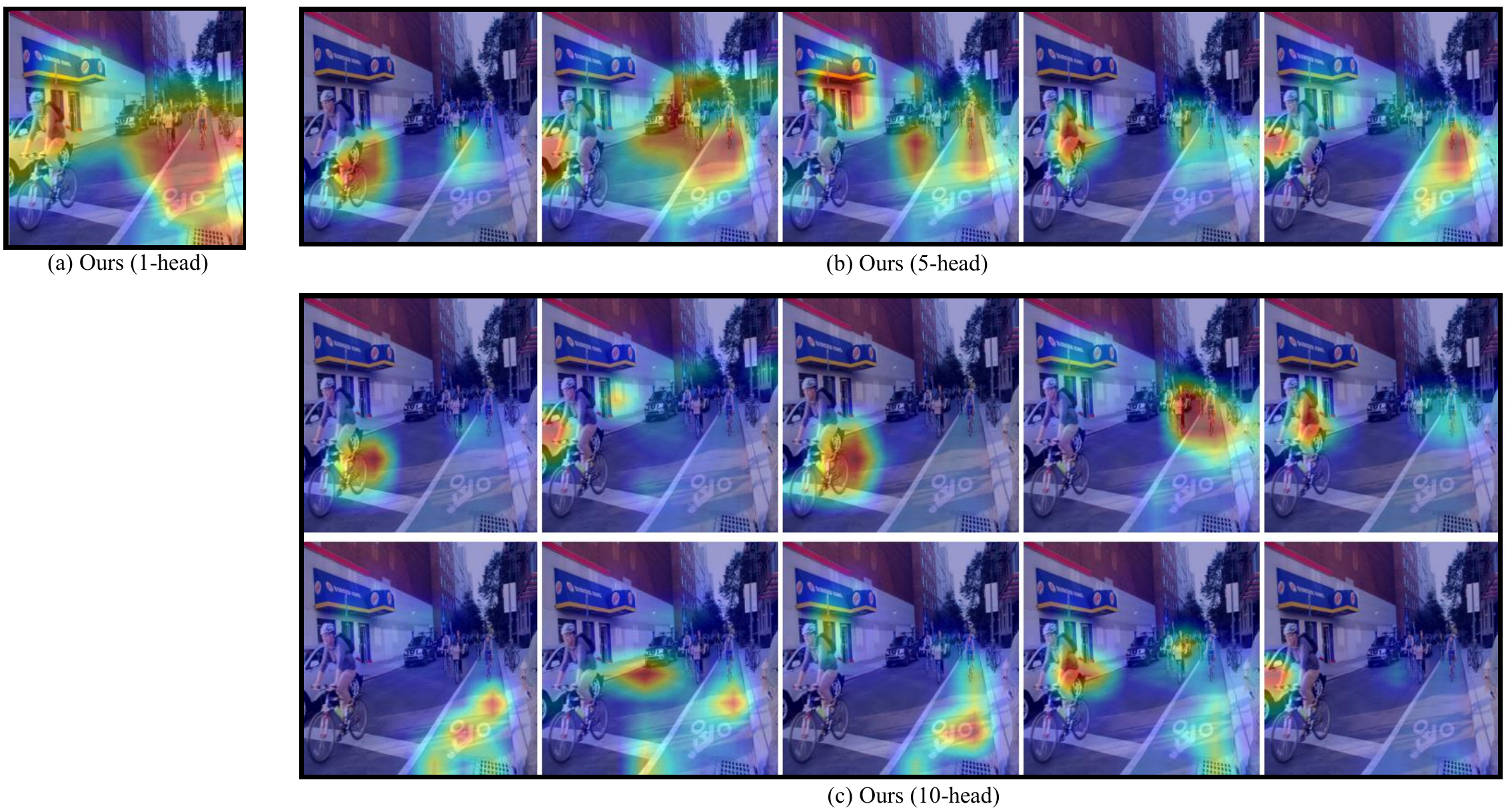}
\end{center}
\vspace{-0.3cm}
   \caption{An example of image attention heatmaps shows the comparison of ours with varying numbers of attention (multi-head). While self-attention with single-head (a) is globally collapsing the street, the crowd, and the person riding the bike in one attention map, the multi-head self-attention (b),(c) show that the,they are attended to separately in multiple attention maps.}
\label{fig:numberofhops1}
\end{figure*}
\begin{figure*}[h]
\begin{center}
\includegraphics[width=0.7\textwidth]{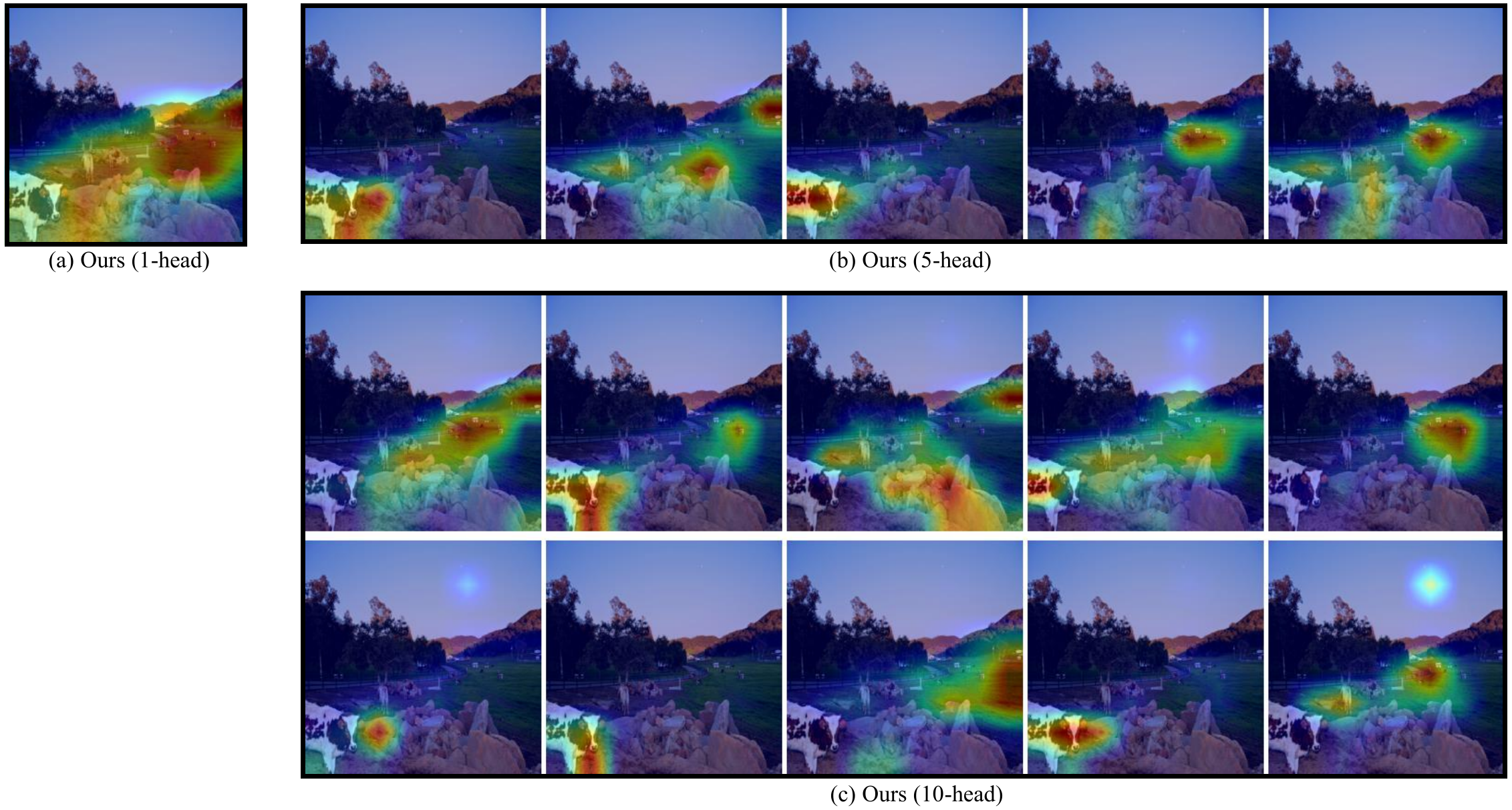}
\end{center}
\vspace{-0.3cm}
   \caption{An example of image attention heatmaps show the comparison of ours with varying numbers of attention (multi-head). While self-attention with single-head (a) is globally collapsing the green field with other objects such as the cow in one attention map, the multi-head self-attention (b),(c) show that the they are attended to separately in multiple attention maps.}
\label{fig:numberofhops2}
\end{figure*}
\subsection{Comparison of Multi-Head Self-Attention and Single-Head Self-Attention}\label{ssec:sub2_ex}
We improve the performance of image-text retrieval tasks by extending our model to multi-head self-attention as shown in Table~\ref{table:1} and Table~\ref{table:2}. To observe the effect of the number of attention weight vectors in each encoder, we experiment by fixing one encoder without self-attention and changing the number of attention maps in the other. Table~\ref{table:4} shows that more attention maps result in better performance and we achieve the state-of-the-art results on image-text retrieval by using 10-head self-attention in both encoders. Also, we observe that applying the multi-head self-attention only to the text encoder is more effective than the image encoder only.\\
\textbf{Visualization.} To observe how the image is encoded into the joint space with our proposed model, we visualize the attention heatmaps learned by the model by varying the number of heads in Figure~\ref{fig:numberofhops1},~\ref{fig:numberofhops2}. Although self-attention with a single-head captures the globally important region of the image, the attention weights are distributed over all important region of the image as shown in Figure~\ref{fig:numberofhops1},~\ref{fig:numberofhops2} (a). On the other hand, the model with 5-head separately provide the information of different semantic regions by generating five attention weight vectors as shown in Figure~\ref{fig:numberofhops1},~\ref{fig:numberofhops2} (b). Comparing ours (10-head) to ours (5-head), we show clearly that the object-level regions in the image are concentrated in each attention heatmap as visualized in Figure~\ref{fig:numberofhops1},~\ref{fig:numberofhops2} (c). 
This result verifies that our model with the multi-head self-attention provides the diverse aspects of semantics in the image to visual-semantic embedding, resulting in better performance than the single-head self-attention. Moreover, we are able to interpret the meaning of each vector in the image embedding using attention heatmaps. For example, as shown in the first attention map in Figure~\ref{fig:numberofhops1} (c), the first row of the embedding matrix ($\bm{F}$) in 10-head means the semantics of the bike in the image and the fourth row of embedding matrix ($\bm{F}$) means the person riding the bike as shown in the fourth attention map.

\subsection{Effect of Diversity Regularization in Loss}\label{ssec:sub3_ex}
\begin{figure*}[t]
\begin{center}
\includegraphics[width=0.9\textwidth]{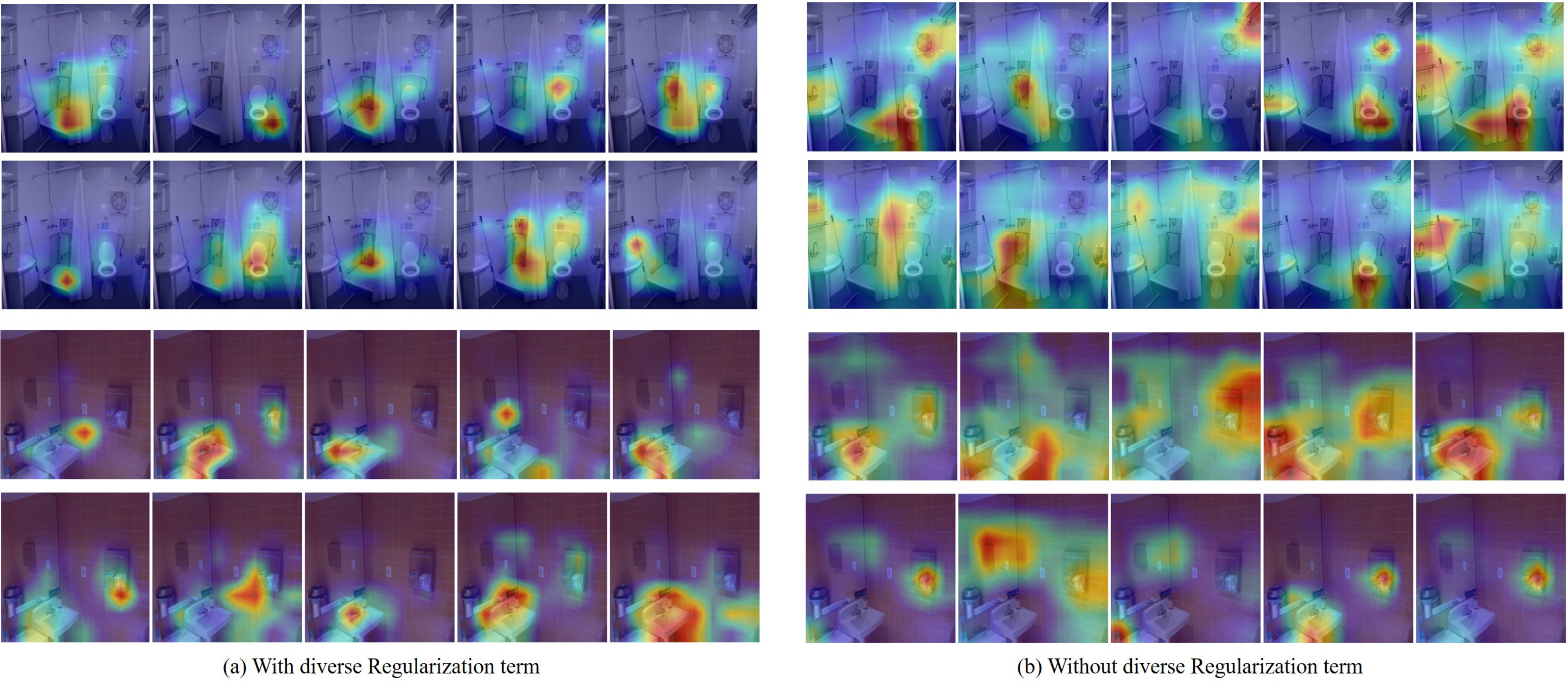}
\end{center}
\vspace{-0.3cm}
\caption{Effect of diversity regularization term. We train our model (10-head) (a) with diverse regularization term in loss (coefficient=0.1) and (b) without diverse regularization term in the loss.}
\label{fig:diversity}
\end{figure*}

In this subsection, we experiment to observe the effect of diversity regularization in the loss. For comparison, we report the average diversity loss value described in equation~\ref{eq:regularization} on the 5k test set and performance of image-text retrieval tasks come from our model (10-head) trained with and without diverse regularization term in Table~\ref{table:5}. These results in Table~\ref{table:5} show that diversity regularization has a great effect on image-text retrieval performance. We conjecture that diversity regularization encourages encoded vectors to contain various semantic meanings in an image, reducing the redundancy of the encoded vectors and attention weight vectors.\\
\textbf{Visualization.} We qualitatively present the multiple attention maps focusing on more distinct local regions by introducing the diversity regularization. As shown in Figure~\ref{fig:diversity}, the diversity regularization term in the loss function makes attention matrix sparse, which encourages different regions to be captured by each attention map. Therefore, the image embedding encodes more distinct components in an image, resulting in an improvement of performance on the image-text retrieval tasks.

\begin{table*}[t]
\centering
\begin{adjustbox}{max width=0.8\textwidth}
\begin{tabular}[b]{r|ccc}
Diversity Regularization&Sentence Retrieval&Image Retrieval&Diversity Loss\\
\hline
W/&\textbf{73.5}&\textbf{59.1}&\textbf{5.59}\\
\hline
W/O&72.3&58.4&7.36\\
\end{tabular}
\end{adjustbox}
\caption{Impact of diversity regularization on performance of retrieval. The lower the diversity loss, the more diverse the local regions focused by the multiple attention maps.}
\label{table:5}
\end{table*}
\section{Conclusion}
We presented an interpretable and powerful multi-head self-attention network for visual-semantic embedding. Different from existing visual semantic embedding models, our proposed network represents visual and textual data in multiple vectors generating multiple attention weights where each weight encodes a different part in the data. Furthermore, we enhance the model with diversity regularization loss to address the redundancy problem where the multiple attention mechanism provides similar representations for the data. Experiments on two benchmark datasets show that our model outperforms state-of-the-art visual semantic embedding models by a large margin. Through the visualization of attention maps, we demonstrate that our proposed method provides various aspects of visual and textual data by encoding semantically distinct parts of the image and the text separately. Notably, our model is able to achieve superior performance without any other resources such as bounding box or image segmentation datasets. Since we achieve the significant improvement by applying the multi-head attention module independently in a modality encoder, we expect that our method can be widely applied for other computer vision tasks such as image classification, visual question answering ,and semantic segmentation.

{\small
\bibliographystyle{ieee}
\bibliography{egpaper_final}
}
\end{document}